\documentclass[11pt]{article}
\usepackage{acl}
\usepackage{times}
\usepackage{latexsym}
\usepackage[T1]{fontenc}
\usepackage[utf8]{inputenc}
\usepackage{microtype}
\usepackage{inconsolata}
\usepackage{graphicx}
\usepackage{booktabs}
\usepackage{amsmath}
\usepackage{amssymb}
\usepackage{amsthm}
\usepackage{enumitem}
\graphicspath{{figures/}}

\newtheorem{proposition}{Proposition}
\newtheorem{definition}{Definition}
\newcommand{\rrp}{\textsc{RRP}}
\newcommand{\ps}{p_{S}}
\newcommand{\qone}{q_{1}}
\newcommand{\qtwo}{q_{2}}
\newcommand{\acc}{\mathrm{acc}}

\title{JevCascade: Readout Stability and Inference-Time Compute Allocation\\
	for Prefill-Only Decision Models}

\author{
	Ran Li\textsuperscript{1} \quad
	Lei Chen\textsuperscript{1,2} \\
	\textsuperscript{1}Hong Kong University of Science and Technology  \\
	\textsuperscript{2}Hong Kong University of Science and Technology(Guangzhou)  \\
	\texttt{rlibb@connect.ust.hk} \quad \texttt{leichen@cse.ust.hk}
}

\begin{document}
	\maketitle
	
	\begin{abstract}
		We present JevCascade, an inference-time compute allocation framework for prefill-only decision models, built on readout stability and zero-label restricted readout prediction. Prefill-only decision models score every candidate in a menu in a single forward pass and never decode, which makes one call one to two orders of magnitude cheaper than a same-scale generative language model. We show that this read-out structure carries a testable property. When an intervention changes only the menu and leaves the input text fixed, the post-intervention accuracy is already determined by the cached first-pass distribution. The restriction predictor restricts the pass-1 probabilities to the menu, renormalizes and reads off the argmax, with no labels and no second pass. Across six model families, ten core datasets (sixteen in total) and two task types, menu-only interventions are predicted to within 4.2 points, exactly for one family. The property is ordinal, not probabilistic, since a probability-level variant of the same estimator errs by 21.0 points, and class-specific, since same-scale generative models err by 1.6 to 15.8 points and degrade as they grow. It makes inference-time compute decidable before deployment. Uniform extra passes buy calibration but almost no accuracy; at matched cost a confidence cascade beats every scheme that re-asks the same model, and curating the menu beats enlarging the model, a 0.8B model on a curated 5-candidate menu reaching 95.4\% on CLINC150 against 80.0\% for a 4B model on the full 150-label menu. Code and data are available at \url{https://github.com/rlisml/jev-cascade}.
	\end{abstract}
	
	\section{Introduction}
	
	A prefill-only decision model reads an input text once and scores every candidate label with an independent head, returning a distribution over a menu of candidates without decoding a single token. This read-out structure makes one call one to two orders of magnitude cheaper than a same-scale generative language model~\citep{cheng2026model,typesafe2026systemone}, so the models serve as drop-in classifiers for intent detection, query routing and content moderation~\citep{deusser2026evaluating,li2026replacing} and as calibrated decision heads inside larger agents~\citep{guo2026just,wu2026reflex}.
	
	\begin{figure}[t]
		\centering
		\includegraphics[width=0.235\textwidth]{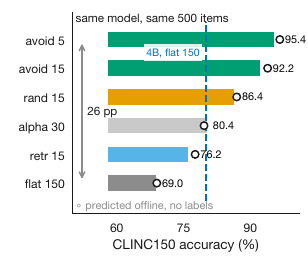}\hfill
		\includegraphics[width=0.235\textwidth]{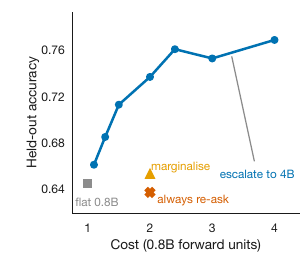}
		\caption{Left: Menu-only changes move CLINC150 accuracy across 26 points; bars = measured, circles = predictions. Right: At matched cost, escalating from kev-0.8B (0.8B forward units) to kev-4B beats re-asking.}
		\label{fig:motivation}
	\end{figure}
	
	For a deployer, the model is only half of the decision. The other half is the menu, the set of candidate labels offered at inference time, and it is built by the deployer rather than fixed by the model, whether by sampling a random subset, retrieving from a front end, grouping labels into a hierarchy, or editing the label table itself. Prior work has changed menus in each of these ways, by pruning or reorganizing the label space~\citep{lu2024mitigating,yang2026semantic,li2026localize} or by making the presentation of options more stable~\citep{pezeshkpour2024large,zheng2023large}. Figure~\ref{fig:motivation} shows that this choice is not a detail. Holding the model and the 500 test items fixed on CLINC150~\citep{larson2019evaluation}, moving from the full 150-label menu to a curated 5-candidate menu raises accuracy from 69.0\% to 95.4\%, a span of 26 points that is wider than the gap between the 0.8B and 4B models on the same data. Changing the menu is a first-order intervention, and a deployer will make it repeatedly.
	
	Evaluating such an intervention is expensive in the currency that matters. A second forward pass is cheap, but scoring the result needs labels, which are not. Existing label-free machinery does not fill the gap. Conformal and selective-prediction methods attach a coverage or abstention guarantee to a fixed predictor~\citep{vishwakarma2024prune,quach2024conformal,kiyani2025decision,noorani2026conformal}, and unsupervised accuracy estimators return one number for a whole dataset~\citep{merdjanovska2026evaluation}. This paper shows that the evaluation can be skipped. When only the menu changes and the input text is held fixed, the post-intervention accuracy is already determined by the cached first-pass distribution. Restrict the pass-1 probabilities to the menu, renormalize, and read off the argmax. We call the estimator the \emph{restriction predictor} and the underlying property \emph{readout stability}. The predictor uses no labels, no second forward pass and no training, and prices each concrete menu separately.
	
	The property is sharp enough to be useful. Across six model families, ten core datasets (sixteen in total) and two task types, menu-only interventions are predicted to within 4.2 accuracy points, and for one family it is exact, because its second pass carries no information the first pass did not already contain. The same property does not hold for same-scale generative language models, which err by 1.6 to 15.8  and get worse as they grow, consistent with the finding that their stated confidence is only loosely tied to their computation~\citep{kadavath2022language,xiong2024can}. The difference follows from the read-out structure. A decision model scores each candidate with a head that reads a shared text representation, so changing the menu only changes which heads are read; a generative model writes the menu into the prompt, so changing the menu changes the text it conditions on.
	
	Two details keep the claim from being a restatement of normalization. First, the property is ordinal, not probabilistic. Predicting accuracy from the pass-1 \emph{probabilities} errs by 21.0 points, while predicting it from the pass-1 \emph{ranking} errs by 1.33 points, so no calibration method recovers the property and no probability-level shortcut replaces the predictor, even though calibration is otherwise the standard remedy for biased confidence~\citep{mao2025calibrating,loginova2025addressing,li2025calibraeval}. Second, the predictor answers a question that unsupervised accuracy estimation does not. It prices a specific intervention rather than the dataset, and label-free constants that estimate one number per dataset err by 18 to 21 points on the same cells.
	
\begin{figure*}[t]
		\centering
		\includegraphics[width=0.95\textwidth]{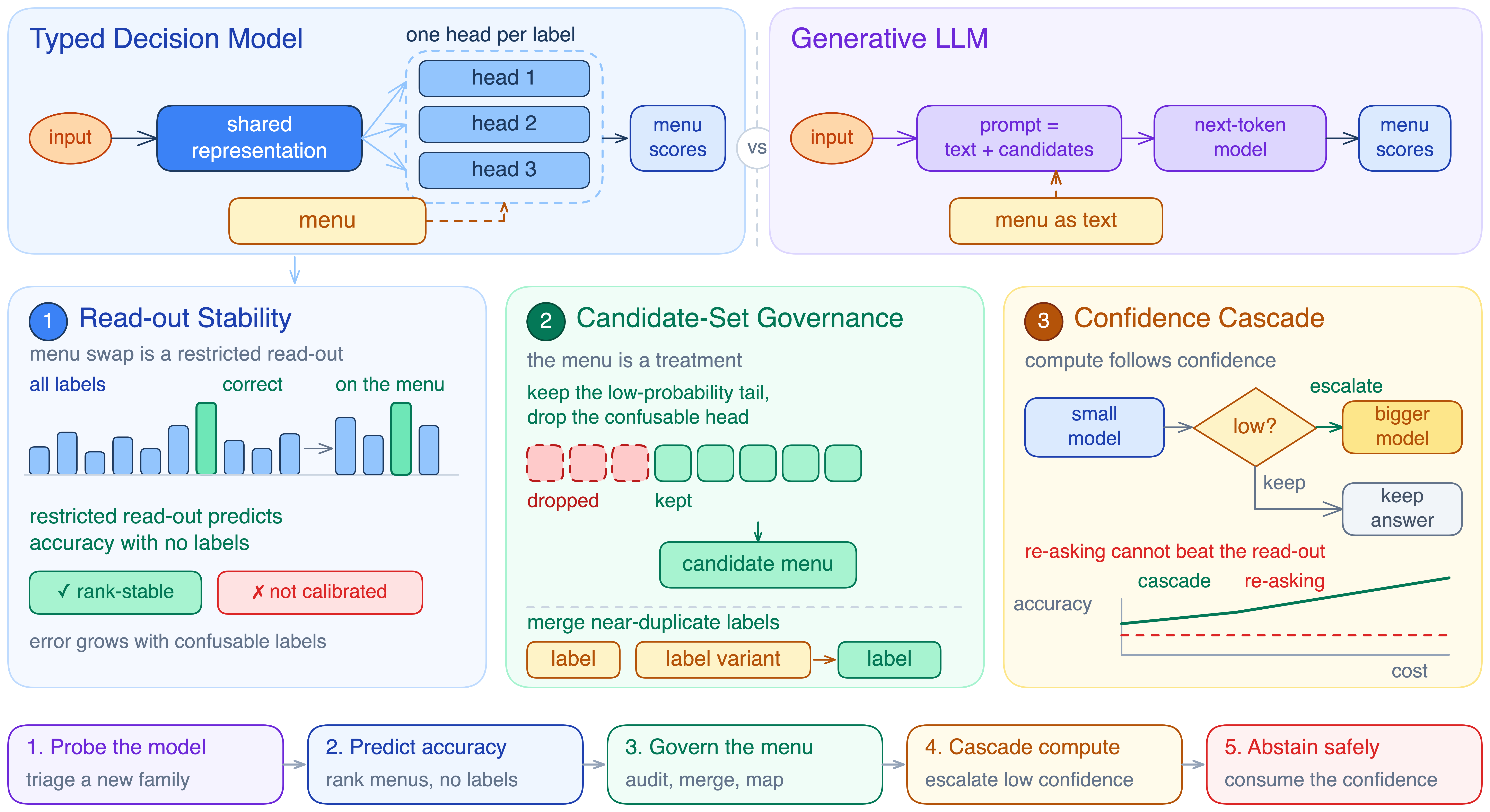}
		\caption{The two read-out structures (top) and the three threads of the paper (bottom).}
		\label{fig:overview}
	\end{figure*}

	The property also changes how inference-time compute should be spent. If every menu-only intervention is free to evaluate, compute is no longer needed to explore menus, and the budget can be routed elsewhere. Figure~\ref{fig:motivation} (right) previews the answer. Uniform extra passes, which are the default way to spend test-time compute in generative models~\citep{wang2022self}, buy calibration but almost no accuracy. At matched cost, escalating the least-confident items to a larger sibling dominates every scheme that re-asks the same model, mirroring the advantage that cascades and routers already show for generative models~\citep{chen2023frugalgpt,dohan2022language,dekoninck2024unified,rabanser2026gatekeeper}. Curating the menu dominates enlarging the model. Figure~\ref{fig:overview} lays out the three threads of the paper, namely the stability property, the predictor that exploits it, and the allocation recipe that follows. We package the three into a deployment framework we call JevCascade.
	We make four contributions.
	\begin{itemize}[wide=0pt, itemsep=2pt, topsep=2pt, parsep=0pt]
		\item \emph{A testable property.} We identify readout stability and show that it holds for menu-only interventions across six model families, ten core datasets (sixteen in total) and two task types, with a worst-case error of 4.2 points and an exact prediction for one family.
		\item \emph{A predictor and a theory of its limits.} We define the restriction predictor, show that it is exact whenever the model is readout-stable, and characterize the two mechanisms behind its residual, one tied to rival confusability and one to menu size. A per-item flip model explains the largest exception within 2 points.
		\item \emph{Class specificity.} We show that same-scale generative language models violate the property, that the violation grows with model size, and that the difference follows from how the two classes read out a menu.
		\item \emph{An allocation recipe.} We turn the property into deployment decisions. Uniform passes buy calibration, confidence cascades outperform re-asking at matched cost across three families, and candidate-set curation beats a larger model.
	\end{itemize}
	
	\section{Related Work}
	
	\paragraph{Uncertainty and selective prediction.}
	Whether a model's own confidence can be trusted is studied. \citet{kadavath2022language} and \citet{xiong2024can} show that language models carry usable uncertainty signals, which conformal methods turn into coverage guarantees~\citep{vishwakarma2024prune,quach2024conformal,kiyani2025decision,noorani2026conformal} and selective-prediction and abstention methods spend on coverage~\citep{lee2024selective,mao2025calibrating,wang2026safer,tjandra2024fine}; \citet{merdjanovska2026evaluation} document how easily such estimates are misread. We instead price a specific menu intervention offline, so a probability-level estimator fails where the ranking-level one succeeds.
	
	\paragraph{Test-time compute, routing and cascades.}
	Self-consistency samples more and majority-votes~\citep{wang2022self}, whereas cascades and routers spend the budget on model choice instead, from FrugalGPT~\citep{chen2023frugalgpt} and language-model cascades~\citep{dohan2022language} through token-level~\citep{gupta2024language} and mixture-of-thought variants~\citep{yue2024large}, learned routing~\citep{ding2024hybrid,chuang2024learning,damani2025learning} and a unified routing-and-cascading treatment~\citep{dekoninck2024unified} to cost- or confidence-tuned variants~\citep{valkanas2026c3po,lee2026confidence,chen2024data,fanconi2026cascaded,rabanser2026gatekeeper}. We add a measurement rather than a router, showing that any re-ask scheme is bounded offline and that a cascade at matched cost outperforms it.
	
	\paragraph{Order sensitivity and label-space reduction.}
	Generative models are sensitive to option order and wording~\citep{pezeshkpour2024large,zheng2023large}, an obstacle in multiple-choice evaluation and judging that appears as primacy effects~\citep{raimondi2025exploiting}, positional bias~\citep{li2025anchored}, blind guessing~\citep{loginova2025addressing} and judge selection bias~\citep{li2025calibraeval}, and that grows with the number of options~\citep{raman2025reasoning,park2025bridging,lin2026overwhelmed}. A separate line reduces the label space through label similarity or semantics~\citep{lu2024mitigating,yang2026semantic,badhe2026silent,li2026localize,ozdemir2026explaining}. Both change a menu but neither prices it; we predict what accuracy the reduced menu will produce.
	
	\paragraph{Decision models.}
	Recent work deploys typed decision models for low-latency serving~\citep{deusser2026evaluating,li2026replacing,cheng2026model}, studies their robustness to prompt injection~\citep{wu2026decision}, and uses their calibrated decisions for alignment checks and selective control~\citep{guo2026just,wu2026reflex}. We characterize a structural property of the read-out they rely on. Faithfulness results for generative models~\citep{turpin2023language,chen2025reasoning} make the contrast plain, since their stated reasons need not track their computation.
	
	\section{Readout Stability and the Restriction Predictor}
	\label{sec:theory}
	
	\subsection{Setup and notation}
	A decision model $M$ takes an input text $x$ and a menu $S$ drawn from a label set $L$, and returns a distribution $q(\cdot \mid x, S)$ over $S$ in one forward pass. We write $C = |S|$ for the menu size, $y(x) \in L$ for the correct label, and $\acc(M,S)$ for the accuracy of the argmax of $q(\cdot \mid x, S)$ averaged over a test set. We call the pass in which $S = L$ the \emph{first pass} and write its distribution as $\qone(\cdot \mid x)$, normalized over $L$. A menu-only intervention replaces $S = L$ with some $S \subseteq L$ while leaving $x$ unchanged, and yields a \emph{second pass} distribution $\qtwo(\cdot \mid x, S)$.
	
	The two model classes differ in how a menu enters the computation. A decision model assigns each label an independent head that reads a shared text representation; the menu selects which heads are read and leaves the representation untouched. A generative model serializes the menu into the prompt, so the menu changes the conditioning text. Section~\ref{sec:evidence} shows that this structural difference is empirically decisive.
	
	\subsection{The restriction predictor}
	Given the cached first-pass distribution, the restriction predictor (\rrp) forms
	\begin{equation}
		\ps(o) = \frac{\qone(o \mid x)}{\sum_{o' \in S} \qone(o' \mid x)},
	\end{equation}
	and predicts the post-intervention accuracy as the fraction of items for which $\arg\max_{o \in S} \ps(o)$ equals $y(x)$. The predictor needs one pass over the data, which is already cached, and  no labels, no second forward pass, no training. It prices each menu separately, so a deployer can rank many candidate menus offline and deploy only the winner.
	
	\subsection{Readout stability}
	The predictor is interesting only to the extent that it matches what the model actually does after the menu changes. The property it relies on is the following.
	
	\begin{definition}[Readout stability]
		\label{def:stability}
		A decision model $M$ is \emph{readout-stable} if, for every $x$ and $S \subseteq L$,
		\begin{equation}
			\label{eq:P}
			\arg\max_{o \in S} \qtwo(o \mid x, S) = \arg\max_{o \in S} \ps(o).
		\end{equation}
	\end{definition}
	
	Under Definition~\ref{def:stability} the predictor is exact by construction. The second pass is correct on $x$ exactly when the restricted distribution ranks $y(x)$ first, so $\acc(M,S) = \Pr[\arg\max_{o \in S} \ps(o) = y(x)]$ holds as an identity, and the residual $\Delta(S)$ is the frequency with which the identity fails. The definition earns its place as a falsifiable hypothesis about a model class rather than as a theorem. Section~\ref{sec:evidence} tests it on 91 interventions, and for one family it holds exactly, so that family's second pass is a lookup into the first and carries no new information. For the others it holds to within a small residual whose structure we characterize next.
	
	\subsection{Where it breaks, and why}
	The residual of \rrp{} is not noise. It is concentrated on items where the first pass is nearly tied, and it has two identifiable drivers.
	
\begin{figure*}[t]
		\centering
		\includegraphics[width=0.21\textwidth]{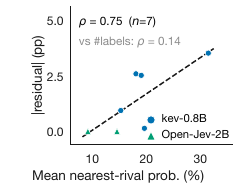}\hfill
		\includegraphics[width=0.21\textwidth]{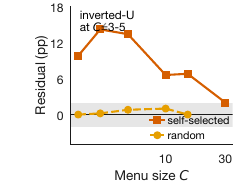}\hfill
		\includegraphics[width=0.21\textwidth]{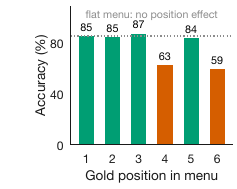}\hfill
		\includegraphics[width=0.21\textwidth]{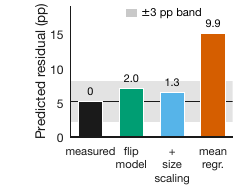}
		\caption{Where the property breaks. From left to right, the panels show residual against confusability, residual against menu size, position on a small menu, and a per-item postdiction of the hardest exception.}
		\label{fig:mechanism}
	\end{figure*}

	\paragraph{Mechanism A (confusability).} A second pass can only overturn a first-pass decision when the top two candidates are close, so the residual should track the probability of the closest rival rather than the number of labels. Empirically it does. Across sources and families the residual correlates with the closest-rival probability at $\rho = 0.75$ and with the label count at $\rho = 0.14$, while the label count is nearly irrelevant (Figure~\ref{fig:mechanism}, first panel). The predictor's error follows how confusable a label space is, not how large it is.
	
	\paragraph{Mechanism B (menu size).} Renormalization rescales the pass-1 probabilities by the mass of the menu, and any fixed positional or wording preference is rescaled with them. The effect shrinks as the menu grows. For self-selected menus the residual falls from $+9.8$ points at two candidates to $+2.0$ at thirty (Figure~\ref{fig:mechanism}, second panel). Small menus are where a residual is most likely.
	
	\paragraph{A per-item flip model.} The two mechanisms act per item, not on the mean. We fit a flip model that predicts whether the second pass is correct from the top-two gap, the rank of the correct label, and whether that label leads the restricted distribution, calibrating on six self-selected menus, and evaluate it on the hardest exception in our grid, an alphabetically ordered 8-candidate menu with a measured residual of $+5.2$ points. The per-item model predicts $+7.15$ points, an error of 1.95 points, against 9.9 points for a mean-level regression on the same variables (Figure~\ref{fig:mechanism}, fourth panel). We treat it as a mechanism-consistent postdiction rather than a general predictor, since its calibration set is small.
	
	\section{Evidence for a Real, Ordinal and Class-Specific Property}
	\label{sec:evidence}
	
	\subsection{Stability across models, datasets and tasks}
\begin{figure*}[t]
		\centering
		\includegraphics[width=0.21\textwidth]{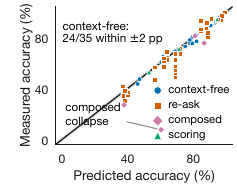}\hfill
		\includegraphics[width=0.21\textwidth]{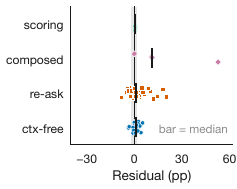}\hfill
		\includegraphics[width=0.21\textwidth]{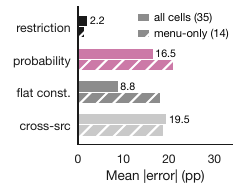}\hfill
		\includegraphics[width=0.21\textwidth]{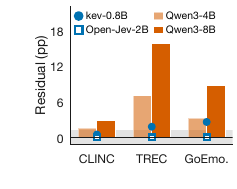}
		\caption{Readout stability. From left to right, the panels show prediction against measurement, residual by intervention family, label-free baselines, and the same estimator applied to generative models.}
		\label{fig:stability}
	\end{figure*}

	We evaluate \rrp{} on 91 menu-only interventions from six decision-model families, kev, Open-Jev, decider, this-that, Mica-4B and Tiny-Jev, all instances of the System One family~\citep{kev2026,openjev2026,marosi2026decider,cheng2026model,mica2026,tinyjev2026,typesafe2026systemone,deusser2026evaluating}, over sixteen datasets spanning intent detection, question typing, multiple-choice knowledge and commonsense, topic classification, emotion and ordered scoring (Appendix~\ref{app:data}). The interventions cover random, avoidance-style, retrieval, alphabetical and self-selected menus, at menu sizes from two to thirty. The first panel of Figure~\ref{fig:stability} plots prediction against measurement. The context-free interventions cluster on the diagonal, with 24 of 35 cells within $\pm 2$ points. Across the full family-by-source grid the error stays within 4.2 points, and the single largest exception, at 5.2 points, is the alphabetically ordered 8-candidate menu discussed.
	
	The second panel aggregates the residuals by intervention family and separates the context-free interventions, whose median is near zero, from re-ask and composed interventions, whose residuals grow because the second pass is no longer a restriction of the first. Table~\ref{tab:family-residual} gives the signed residual for each model family and menu type. The Open-Jev family is exact on every context-free cell, because its second pass is a restriction of its first. The other families show small residuals whose sign is stable within a family, a consistent error.
	
	\begin{table}[b]
\centering
\footnotesize
\setlength{\tabcolsep}{3pt}
\caption{Residual (pp) by model family and menu type. Bold marks an exact prediction; positive entries are overestimates.}
\begin{tabular}{@{}lrrrrr@{}}
\toprule
\textbf{Family} & \textbf{rand 15} & \textbf{avoid 15} & \textbf{self clean} & \textbf{self verify} & \textbf{self 2} \\
\midrule
kev-0.8B & $+0.5$ & $+1.7$ & $+6.8$ & $-0.8$ & $+9.8$ \\
OJ-2B & $\mathbf{+0.0}$ & $\mathbf{+0.0}$ & $\mathbf{+0.0}$ & $+9.4$ & $-0.4$ \\
dec-0.8B & $-0.8$ & $-0.6$ & $-3.4$ & $-3.0$ & $-2.8$ \\
dec-2B & $-0.4$ & $-0.4$ & $-2.0$ & $-2.6$ & $-3.6$ \\
this-that & $+0.6$ & $+3.2$ & $-0.4$ & $-1.8$ & $-8.2$ \\
Mica-4B & $+2.2$ & $+0.2$ & $+8.2$ & $+8.8$ & $+5.2$ \\
Tiny-Jev & $+1.4$ & $+1.4$ & $+1.0$ & $+4.8$ & $+0.8$ \\
\bottomrule
\end{tabular}
\label{tab:family-residual}
\end{table}

	\begin{table}[b]
\centering
\footnotesize
\setlength{\tabcolsep}{4pt}
\caption{Three-tier kev cascade (0.8B$\rightarrow$4B$\rightarrow$9B). Costs are in 0.8B forward units; match-9B is the cheapest cascade reaching 9B accuracy. Bold marks the cascade optimum.}
\begin{tabular}{@{}lrrrrrr@{}}
\toprule
\textbf{Source} & \textbf{0.8B} & \textbf{4B} & \textbf{9B} & \textbf{match-9B} & \textbf{best} & \textbf{best cost} \\
\midrule
AG News & 88.8 & 89.8 & 89.2 & 1.04 & \textbf{91.0} & 1.63 \\
DBpedia & 99.2 & 99.4 & 99.4 & n/a & 99.2 & 1.02 \\
SciQ & 95.9 & 99.0 & 99.2 & 1.80 & \textbf{99.4} & 2.07 \\
ARC & 61.6 & 92.2 & 94.4 & 5.88 & \textbf{94.8} & 6.82 \\
CSQA & 58.0 & 77.6 & 78.2 & 5.43 & \textbf{78.8} & 5.89 \\
OBQA & 50.6 & 84.0 & 85.8 & 6.04 & \textbf{87.2} & 6.83 \\
MMLU & 41.5 & 72.4 & 74.4 & 6.25 & \textbf{75.4} & 6.59 \\
\bottomrule
\end{tabular}
\label{tab:cascade-cost}
\end{table}

	The property extends beyond classification. On SST-5, a five-level ordered scoring task, we treat a menu as a subset of grade levels and predict the accuracy of each subset by restricting the pass-1 distribution and renormalizing. All four subsets are predicted to within 0.8 points, so the property is not tied to a discrete label set.
	
	\subsection{The property is ordinal, not probabilistic}
	A natural objection is that \rrp{} is a renormalization, and that any reasonable estimator of the pass-1 distribution would do. It would not. The predictor works because the \emph{ranking} of the pass-1 distribution survives a menu change, not because the probabilities are accurate. Predicting accuracy from the mean pass-1 probability of the correct label errs by 21.0 points on menu-only interventions, against 1.33 points for the argmax version (Figure~\ref{fig:stability}, third panel). The pass-1 probabilities are systematically too high after a menu change, but which candidate they rank first is almost unchanged, so the property cannot be recovered by calibration and cannot be replaced by a probability-level shortcut.
	
	The same panel rules out the simpler label-free baselines. Assuming the accuracy is unchanged by the menu errs by 18.0 points, and a cross-source empirical constant errs by 18.7 points, both roughly fifteen times the error of \rrp{}. The predictor wins on 34 of 35 cells against the probability-level estimator and on 34 of 35 against the cross-source constant, so its advantage is not within one dataset.
	
	\subsection{The property is class-specific}
	If readout stability followed from renormalization alone, a generative model would have it too. It does not. We give Qwen3-4B-Instruct and Qwen3-8B the same 15-candidate menus as the decision models, read their option scores from the next-token distribution, and predict a second, differently seeded menu from the first. The decision models err by about 0.5 points, while the generative models err by 1.6 to 15.8 points, and the larger model errs more (Figure~\ref{fig:stability}, fourth panel). The failure follows the read-out structure. A generative model conditions on a different text once the menu changes, so its second pass is not a restriction of its first.
	
	The contrast sharpens the contribution. The predictor is not a better estimator of an underlying quantity that both classes share; it is an estimator of a quantity that only one class has. The same comparison also shows that spending sixteen samples on self-consistency for the generative model leaves its accuracy within 0.4 points of a single option-scoring pass, which is the generative analogue of the uniform-pass result we report next.
	
	\subsection{Ablations}
	We isolate the factors that could drive the residual, on CLINC150 with kev-0.8B unless stated otherwise.
	
	\paragraph{Menu size} changes the magnitude of the residual but not its sign. For self-selected menus the residual peaks at intermediate sizes, reaching $+14.2$ points at three candidates, and shrinks as the menu grows to $+2.0$ at thirty, while random menus of the same sizes stay within $\pm 1$ point, because a self-selected menu keeps the labels the model already finds confusable while a random one dilutes them. Varying only the construction moves accuracy from $0.622$ (self-selected 15) to $0.954$ (avoidance 5), with the predictor's offline ranking matching the measured one. Option order matters only on small menus. On a self-selected 15-candidate menu, putting the correct label first drops accuracy from $0.622$ to $0.426$, whereas on a random 15-candidate menu the same manipulation moves it only from $0.862$ to $0.832$; at 150 labels the permutation flip rate is under 6\%.
	
	\paragraph{Wording.} Holding the menu token-for-token fixed and changing only the question moves Open-Jev-2B accuracy from $0.454$ to $0.714$, a span of 26 points, while shuffling the same menu changes it by at most $0.6$ points. The direction is family-specific. The comparative phrasing that costs Open-Jev-2B 9.4 points is the phrasing that repairs kev-0.8B, lifting it from $0.622$ to $0.698$.
	
	\paragraph{Compute and scale.} Raising the number of order-permuted passes from $1$ to $32$ changes accuracy by at most 3.83 points on any source, with a pooled gain of 1.2, while the expected calibration error falls by 67\% as early as $K = 2$. The residual is not an artifact of small models either. On the same self-selected 15-candidate menu it is $+6.8$ points at 0.8B, $+5.4$ at 4B and $+5.6$ at 9B, and the generative contrast widens with scale.
	
	\section{Allocating Decision-Time Compute}
	\label{sec:compute}
	
	If menu-only interventions are free to evaluate, the budget that would have gone into exploring menus is free to go elsewhere. This section asks where, and answers in three steps. Uniform passes are the wrong default, escalation is the right one, and the menu itself is the highest-value object.
	
	\subsection{Uniform extra passes cannot buy accuracy}
	The generative default for spending test-time compute is to sample more and aggregate. The decision-model analogue is to shuffle the option order, run $K$ passes, and average the resulting distributions with confidence weights. The first panel of Figure~\ref{fig:compute} reports the accuracy gain at $K = 16$ across seven sources. Only MMLU moves significantly, by 3.83 points, and the pooled gain is 1.2 points. A prefill-only read-out has no path diversity, so a pass is a fixed posterior and permuting the menu only adds noise, which averaging removes.
	
	What the passes buy is calibration. The expected calibration error of the averaged distribution falls from $0.09$ at $K = 1$ to $0.03$ at $K = 2$, a 67\% reduction, and then plateaus (Figure~\ref{fig:compute}, second panel). That gain belongs downstream, in abstention and risk control, not in accuracy. Thirty-two passes of the 0.8B model reach 46.0\% on MMLU, while a single pass of the 4B model reaches 72.4\% (Figure~\ref{fig:compute}, third panel). The mode of the posterior is the ceiling of accuracy and averaging cannot move it.
	
	\subsection{Confidence cascades outperform re-asking}
	Re-asking the same model on a smaller menu is the other natural way to spend a second pass, and Section~\ref{sec:theory} shows why it cannot work. Under readout stability the second pass adds no information, so the accuracy of any re-ask scheme is bounded by the \rrp{} prediction. A larger sibling can exceed that bound, because it brings new information. The bound therefore predicts that escalation should outperform re-asking whenever the sibling's flat accuracy exceeds the restricted prediction.
	
	\begin{proposition}[Cascade dominance]
		\label{prop:cascade}
		Suppose $M_s$ is readout-stable. Then the accuracy of any scheme that re-asks $M_s$ on a submenu is at most the restriction prediction. If a sibling $M_b$ has flat accuracy above that prediction, escalating the least-confident fraction $r$ of items to $M_b$ attains accuracy above the bound at expected cost $1 + (u_b - 1) r$, where $u_b$ is the cost of $M_b$ in units of $M_s$. The cascade therefore outperforms re-asking at every cost point at which the bound is exceeded.
	\end{proposition}
	
	On 250 held-out CLINC150 items, the cascade escalates the least-confident items from kev-0.8B to kev-4B using a threshold set on a training split. At matched cost it outperforms every re-ask scheme. It reaches 0.684 at 1.28 units against 0.668 for the best re-ask protocol at 1.44 units, and 0.760 at 2.40 units against 0.652 for order marginalization at 2.00 units. Table~\ref{tab:cascade-cost} reports the same comparison per dataset. Across seven sources the cascade reaches the 9B single-shot accuracy at 1.0 to 6.3 forward units instead of 11, and on six of the seven it exceeds that accuracy outright.
	
\begin{figure*}[t]
		\centering
		\includegraphics[width=0.21\textwidth]{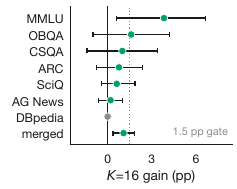}\hfill
		\includegraphics[width=0.21\textwidth]{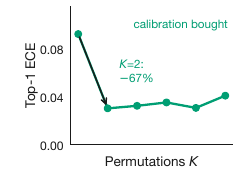}\hfill
		\includegraphics[width=0.21\textwidth]{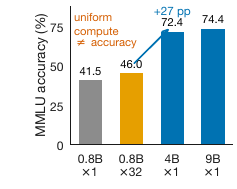}\hfill
		\includegraphics[width=0.21\textwidth]{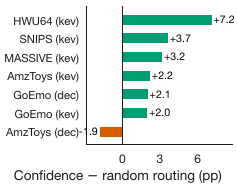}
		\caption{The economics of decision-time compute. From left to right, the panels show the accuracy gain from sixteen order-permuted passes, calibration error against $K$, the equal-budget anchor, and gate quality across tasks.}
		\label{fig:compute}
	\end{figure*}

	The value of a cascade is the product of the gap between the two tiers and the quality of the gate. The gate matters because confidence need not identify the items a larger model would fix. Across five additional sources the kev gate beats random routing on all five, by 2.0 to 7.2 points (Figure~\ref{fig:compute}). The decider family is less reliable. Its gate does not beat random routing on every source, which is why we recommend validating the gate per model and dataset before deploying a cascade.
	
	\begin{figure*}[t]
		\centering
		\includegraphics[width=0.21\textwidth]{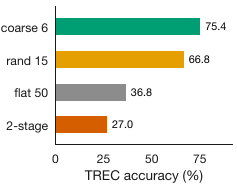}\hfill
		\includegraphics[width=0.21\textwidth]{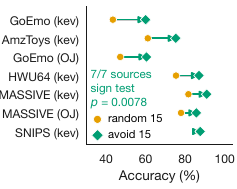}\hfill
		\includegraphics[width=0.21\textwidth]{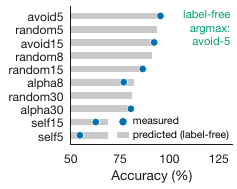}\hfill
		\includegraphics[width=0.21\textwidth]{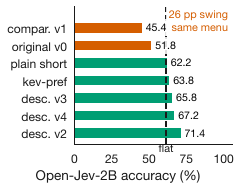}
		\caption{Candidate-set governance and presentation. From left to right, the panels show the TREC hierarchy collapse, avoidance against random menus, offline menu search, and nine phrasings of one menu.}
		\label{fig:governance}
	\end{figure*}
	
	\subsection{Candidate curation beats a larger model}
	The largest single lever is the menu itself. Because a random menu is easy for the model to reject, while a retrieval or self-selected menu pulls the most confusable labels into the candidate set, the accuracy of a wide label space is governed by the density of near-duplicate distractors rather than by the number of labels. On CLINC150 a 0.8B model on a random 15-candidate menu reaches 86.4\%, above a 4B model on the full 150-label menu at 80.0\%. Semantic hierarchies fail for the same reason. Grouping 50 TREC fine classes into a two-stage decision pushes the model into the hardest within-group distinctions and drops accuracy to 0.270, below the flat 50-label menu at 0.368 (Figure~\ref{fig:governance}, first panel).
	
	\begin{proposition}[Random narrowing]
		\label{prop:narrowing}
		Let $\pi$ be the fraction of items on which the first pass does not rank the correct label first, let $\lambda$ be the fraction of items on which every randomly sampled distractor is ranked below the correct label, and let $\alpha$ be the accuracy on those items after narrowing. Then
		\begin{equation}
			\acc(\text{random-}C) \approx (1-\pi) + \pi \lambda \alpha .
		\end{equation}
	\end{proposition}
	
	On CLINC150 the measured values $\pi$ = $\frac{155}{500}$, $\lambda = \frac{93}{155}$ and $\alpha = 0.935$ give 0.864, matching the measured 0.864, so the gain of a random menu is accounted for by which distractors enter the menu.
	
	The predictor turns menu construction into an offline search. Ranking ten candidate-set strategies from cached first-pass distributions, with no second pass and no labels, recovers the measured order and selects the best strategy. The predicted optimum is 0.966 and the measured value is 0.954, the highest in the grid (Figure~\ref{fig:governance}, third panel). Avoidance-style menus, which sample from the low-probability region to keep confusable labels out, are never worse than random menus and are higher on every one of six source-and-model combinations, by 0.6 to 16.5 points (one-sided sign test, $p = 0.008$; Figure~\ref{fig:governance}, second panel). A deployer can therefore choose the menu before running the model at all.
	
	\subsection{Presentation effects are family-specific}
	A menu is delivered through a wording, and the wording is not neutral. Holding a 15-candidate menu token-for-token fixed and changing only the question, accuracy on Open-Jev-2B spans 26 points, from 0.454 for a comparative phrasing to 0.714 for a descriptive one, against 0.614 for the original phrasing (Figure~\ref{fig:governance}, fourth panel). The effect is not a property of the menu text or the option order, since shuffling changes accuracy by at most 0.6 points. It is a property of the family, and it reverses across families. The same comparative re-ask that leaves kev unchanged at $-0.8$ points costs Open-Jev-2B 9.4 points. Order and anchoring move accuracy up to 26 points on the same self-selected menu (Figure~\ref{fig:appendix-results}, third panel).
	
	The practical consequence is that a new family must be scanned before it is deployed, and the scan is cheap. Four configurations of a single dataset, about ten GPU-minutes, reveal whether re-asking carries information, which direction the wording pushes, and whether the restriction prediction is exact for the family. We package this as a probe battery and place it first in the deployment pipeline.
	
	\section{From Prediction to Practice}
	\label{sec:practice}
	
	The result is a deployment recipe in which every stage is decided before it is run, using the cached first-pass distribution as the only input.
	
	\paragraph{Curate before you escalate.} Because the menu is the largest lever, it should be settled first. When the gold label is available offline, avoidance-style menus are the best construction; when the menu is built by a retrieval front end that does not guarantee the gold label, the system reduces to a single recall threshold. Deployment accuracy is approximately the recall of the front end times the accuracy on items it retrieves, so the menu beats the full label set whenever recall exceeds $r^\ast = \text{flat accuracy} / \text{accuracy given a hit}$. For random menus $r^\ast$ lies between 0.65 and 1.00 and the system is deployable, whereas for avoidance menus on the kev family it exceeds one on several sources, because avoidance sampling and the position of the gold label are negatively correlated.
	
	\paragraph{Repair the label table by mapping, not by re-running.} When the confusion is caused by near-duplicate labels, the fix is to merge them, and the merge should be deployed by mapping the cached pass-1 answers into the merged taxonomy rather than by re-running the model. On 2,000 held-out CLINC150 items, mapping gains 5.2 points over the original label table, while re-deciding in the merged space loses 4.2 points relative to mapping (Figure~\ref{fig:appendix-results}, fourth panel). The predictor anticipates the gain. Its offline estimate of the mapped accuracy is 70.6 against a measured 71.4.
	
	\paragraph{Escalate, then abstain.} With the menu settled, the remaining budget goes to escalation and the confidence the cascade produces is worth calibrating for downstream consumption. Two marginalization passes cut the calibration error by two thirds, which is the right input for abstention and risk control.
	
	The pipeline is deliberately cheap. A probe battery triages a new family in about ten GPU-minutes, the offline search ranks menu strategies in seconds, and the cascade threshold is a quantile of the confidence distribution and needs no labels (Figure~\ref{fig:pipeline}).
	
	\section{Conclusion}
	
	Prefill-only decision models are usually discussed as a cheaper way to do what a generative model does. We have argued that they are also a different kind of object. Their read-out structure makes menu-only interventions predictable from a single cached pass, exactly for one family and to within a few points for the rest, while same-scale generative models do not have the property and degrade as they grow. The property is ordinal, so calibration cannot replace it, and it is per-intervention, so aggregate accuracy estimation cannot either. It makes deployment decidable in advance. Menus can be chosen, label tables repaired and compute allocated before anything is run, and the right allocation is the menu and the escalation, not the extra pass.
	
	\section*{Limitations}
	
	Our study is inference-time. Training-time interventions are outside its scope. The restriction predictor is exact for one family and tight for the others. On emotion and product label spaces, where near-duplicate labels are dense, the residual has a fixed positive sign of 2.5 to 4.9 points, so the prediction errs on the conservative side there. We present the per-item flip model as a mechanism-consistent postdiction, and extending it to a general predictor would call for a larger calibration grid, since the residual is not monotone in menu size. The presentation effects we document are localized and reversible across families; their training-side cause is not readable from families whose training data is not public, so we characterize them at the level of the deployed model. Our deployment thresholds assume a retrieval front end whose recall can be measured, which covers the common case of a menu built by a retriever.
	
	\bibliography{jevref}
	
	\appendix
	
	\section{Datasets and Models}
	\label{app:data}
	The six decision-model families are kev~\citep{kev2026} (0.8B/4B/9B), Open-Jev~\citep{openjev2026} (2B/9B), decider~\citep{marosi2026decider} (0.8B/2B), this-that~\citep{cheng2026model}, Mica-4B~\citep{mica2026} and Tiny-Jev~\citep{tinyjev2026}, all instances of the System One family~\citep{typesafe2026systemone,deusser2026evaluating}. The sixteen datasets cover intent detection~\citep{larson2019evaluation,casanueva2020efficient,fitzgerald2023massive,liu2021benchmarking,coucke2018snips}, question typing~\citep{li2002learning}, knowledge and commonsense multiple choice~\citep{hendrycks2020measuring,clark2018think,mihaylov2018can,talmor2019commonsenseqa,welbl2017crowdsourcing}, topic classification~\citep{zhang2015character,auer2007dbpedia}, emotion~\citep{saravia2018carer,demszky2020goemotions} and ordered scoring~\citep{socher2013recursive}.
	
	\section{Additional Results}
	
	\begin{table}[t]
\centering
\footnotesize
\setlength{\tabcolsep}{4pt}
\caption{Residual typology.}
\begin{tabular}{@{}p{2.3cm}p{1.3cm}p{3.5cm}@{}}
\toprule
Residual type & Size (pp) & Mechanism \\
\midrule
pure restriction & $|\Delta| \leq 2$ & context-free menu changes; (P) holds \\
re-ask anchoring & $+6$ to $+20$ & presentation artefact; removable \\
re-ask narrowing & $-3$ to $-4$ & sign flips with label-space confusability \\
de-biasing excess & $-1$ to $-4$ & permutation marginalisation averages out order noise \\
composed collapse & $+11$ to $+53$ & two-stage decision multiplies the pass-1 bias \\
calibration drag & $-9$ to $+3$ & cascade inherits the raw confidence bias \\
serial position & $\approx +5$ @ $C{=}8$ & position prior re-amplified on small menus \\
rank-order collapse & $\approx -20$ & menu ordered by pass-1 rank \\
forced choice & $-31$ to $+9.8$ & sign set by the strongest rival's confusability \\
\bottomrule
\end{tabular}
\label{tab:typology}
\end{table}

	\begin{table}[t]
\centering
\small
\caption{Same-budget comparison on 250 held-out CLINC150 items; costs are in 0.8B forward units.}
\begin{tabular}{lrr}
\toprule
Strategy & Cost (u) & Accuracy \\
\midrule
flat, no re-ask & 1.00 & 0.644 \\
always re-ask (verify) & 2.00 & 0.636 \\
marginalisation $K{=}2$ & 2.00 & 0.652 \\
SARP gate (verify) & 1.44 & 0.668 \\
gate, standard wording & 1.29 & 0.664 \\
\midrule
cascade, escalate 5.6\% & 1.28 & 0.684 \\
cascade, escalate 10.0\% & 1.50 & 0.712 \\
cascade, escalate 28.0\% & 2.40 & 0.760 \\
\bottomrule
\end{tabular}
\label{tab:budget}
\end{table}

	\begin{table}[t]
\centering
\footnotesize
\setlength{\tabcolsep}{3pt}
\caption{Deployment with a menu that excludes the gold label; the system beats flat when recall is at least $r^\ast$.}
\resizebox{\columnwidth}{!}{%
\begin{tabular}{@{}lllrrrrc@{}}
\toprule
Source & Model & Menu & hit rate & acc$|$hit & $r^\ast$ & verdict \\
\midrule
GoEmotions & jev2b & avoid 15 & 0.322 & 0.426 & 0.90 & deployable \\
GoEmotions & jev2b & random 15 & 0.553 & 0.453 & 0.85 & deployable \\
MASSIVE & jev2b & avoid 15 & 0.091 & 0.713 & 0.79 & deployable \\
MASSIVE & jev2b & random 15 & 0.269 & 0.775 & 0.73 & deployable \\
GoEmotions & kev0.8b & avoid 15 & 0.280 & 0.368 & 1.01 & never beats flat \\
GoEmotions & kev0.8b & random 15 & 0.553 & 0.441 & 0.85 & deployable \\
HWU64 & kev0.8b & avoid 15 & 0.050 & 0.463 & 1.34 & never beats flat \\
HWU64 & kev0.8b & random 15 & 0.210 & 0.704 & 0.88 & deployable \\
MASSIVE & kev0.8b & avoid 15 & 0.046 & 0.674 & 0.94 & deployable \\
MASSIVE & kev0.8b & random 15 & 0.269 & 0.815 & 0.78 & deployable \\
SNIPS & kev0.8b & avoid 15 & 1.000 & 0.869 & 1.01 & never beats flat \\
SNIPS & kev0.8b & random 15 & 1.000 & 0.875 & 1.00 & never beats flat \\
\bottomrule
\end{tabular}}
\label{tab:deployment}
\end{table}

	\begin{table}[t]
\centering
\footnotesize
\setlength{\tabcolsep}{4pt}
\caption{Label-table governance on 2{,}000 held-out CLINC150 items.}
\begin{tabular}{@{}lrrr@{}}
\toprule
Label-table branch & re-decide & map (no re-run) & prediction \\
\midrule
auto-merge & 67.20 & 71.40 & 70.60 \\
oracle merge & 68.60 & 69.20 & 69.35 \\
random control & 61.98 & 68.05 & 67.53 \\
\midrule
flat 150 baseline & \multicolumn{3}{c}{66.25} \\
\bottomrule
\end{tabular}
\label{tab:governance}
\end{table}

	\begin{figure}[t]
		\centering
		\includegraphics[width=\columnwidth]{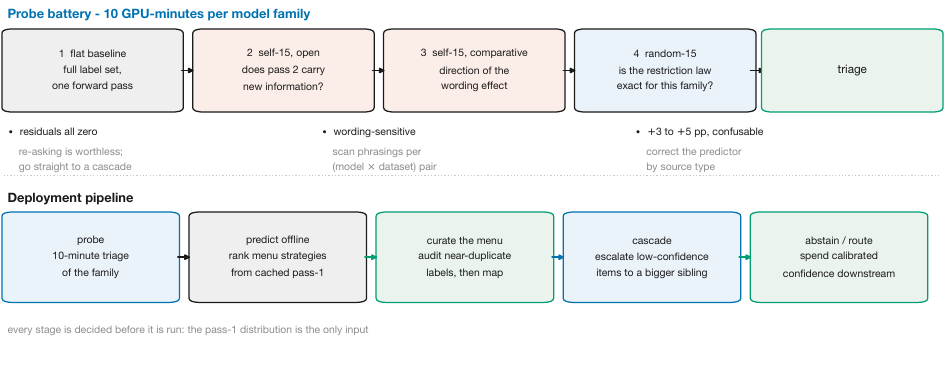}
		\caption{The deployment pipeline.}
		\label{fig:pipeline}
	\end{figure}
	
	\begin{figure*}[t]
		\centering
		\includegraphics[width=0.21\textwidth]{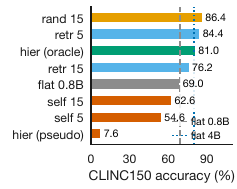}\hfill
		\includegraphics[width=0.21\textwidth]{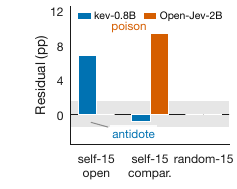}\hfill
		\includegraphics[width=0.21\textwidth]{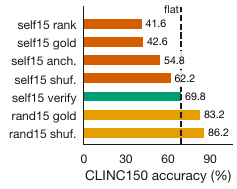}\hfill
		\includegraphics[width=0.21\textwidth]{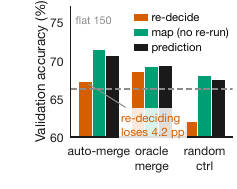}
		\caption{Additional results. From left to right, the panels show CLINC150 menu construction, the cross-family sign flip, the presentation ladder, and label-table repair.}
		\label{fig:appendix-results}
	\end{figure*}
	
	\section{Proofs}
	
	\paragraph{Proposition~\ref{prop:narrowing} (random narrowing).} Items the first pass already gets right stay right, which accounts for the $(1-\pi)$ term. An item the first pass gets wrong is recovered only when no distractor outranks the correct label, an event of probability $\lambda$, and on those items the narrowed menu is correct with probability $\alpha$, giving the $\pi\lambda\alpha$ term. On CLINC150 the measured values $\pi = 155/500$, $\lambda = 93/155$ and $\alpha = 0.935$ give 0.864, matching the measured 0.864.
	
	\paragraph{Proposition~\ref{prop:cascade} (cascade dominance).} Under readout stability the second pass of $M_s$ contains no information beyond its first, so the accuracy of any scheme that re-asks $M_s$ on a submenu is at most the restricted prediction. Escalating the least-confident fraction $r$ of items to a sibling $M_b$ whose flat accuracy exceeds that bound attains accuracy above the bound at expected cost $1 + (u_b - 1) r$. The cascade therefore outperforms re-asking at every cost point at which the bound is exceeded, which is what we measure on three families.
	
	\section{LLM usage}
	\label{app:llm}
	Large-language-model tools were used for editing and formatting support during manuscript preparation.
	
\end{document}